# PARTICLE SWARM OPTIMIZATION FOR ONLINE SPARSE STREAMING FEATURE SELECTION UNDER UNCERTAINTY

*Ruiyang Xu[1]*

[1] *School of Computer Science and Technology, Chongqing University of Posts and Telecommunications, and also with the Chongqing Institute of Green and Intelligent Technology, Chinese Academy of Sciences, Chongqing, China*

**Abstract**

In real-world applications involving high-dimensional streaming data, online streaming feature selection (OSFS) is widely adopted. Yet, practical deployments frequently face data incompleteness due to sensor failures or technical constraints. While online sparse streaming feature selection (OS²FS) mitigates this issue via latent factor analysis-based imputation, existing methods struggle with uncertain feature-label correlations, leading to inflexible models and degraded performance. To address these gaps, this work proposes POS²FS—an uncertainty-aware online sparse streaming feature selection framework enhanced by particle swarm optimization (PSO). The approach introduces: 1) PSO-driven supervision to reduce uncertainty in feature-label relationships; 2) Three-way decision theory to manage feature fuzziness in supervised learning. Rigorous testing on six real-world datasets confirms POS²FS outperforms conventional OSFS and OS²FS techniques, delivering higher accuracy through more robust feature subset selection.

## 1 Introduction

As a fundamental preprocessing step in machine learning, feature selection aims to identify the most informative subset of features while preserving discriminative power in high-dimensional data [1]. This task grows increasingly complex in dynamic streaming environments, where features arrive sequentially within potentially infinite feature spaces [2]. Online streaming feature selection (OSFS) has gained prominence as a scalable solution, enabling incremental evaluation of feature relevance and redundancy under such conditions. However, real-world deployments often encounter data incompleteness due to hardware failures [3] or system limitations [4]. To mitigate this issue, Wu et al. [5] proposed LOSSA, an online sparse streaming feature selection (OS²FS) framework leveraging latent factor analysis (LFA) for missing value imputation. While effective, this method introduces estimation inaccuracies, leading to uncertainty in feature-label associations—a critical limitation in sparse data scenarios.

Traditional feature evaluation methods typically focus solely on complete streaming features, failing to account for the uncertainty caused by missing data. This limitation becomes especially critical considering that feature selection constitutes an NP-hard discrete optimization problem with binary decision variables. Evolutionary computation (EC) techniques have proven particularly valuable in solving such complex optimization challenges. Within the EC domain, particle swarm optimization (PSO) has gained prominence for feature selection tasks owing to two key characteristics: (1) its exceptional global search ability, and (2) relatively straightforward implementation [6]. These advantages have made PSO-based methods particularly prevalent in feature selection applications.

The feature selection process naturally encounters ambiguity due to ill-defined boundaries between features' relevance. This complexity intensifies when considering moderately relevant features - those that may not exhibit strong individual significance but collectively contribute to improved discriminative power in OS²FS systems [7]. These circumstances generate an indeterminate classification zone where conventional binary selection methods fall short. To resolve this, three-way decision (3WD) theory provides a robust solution by incorporating a deferred decision category, allowing for additional evidence gathering before making final feature selections.

This study proposes a novel particle swarm optimization for online sparse streaming feature selection under uncertainty (POS²FS) through two key mechanisms:

a) A novel PSO-based optimization framework that employs fitness-driven particle dynamics to refine feature correlation assessment. This adaptive search strategy demonstrates measurable improvements in feature selection precision compared to conventional methods.

b) A redesigned three-way decision system tailored for ambiguous feature selection contexts. This enhancement specifically targets uncertainty reduction in supervised OS²FS implementations, providing more robust classification outcomes.

## 2 Related Work

Recent developments in dynamic feature selection have seen notable progress through online streaming feature selection (OSFS) approaches. These methods excel at continuously filtering out non-informative and correlated features in data streams. Building upon this foundation, researchers have developed enhanced variants such as the OSFS and fast-OSFS architectures [8], which implement distinct phases for relevance determination and redundancy elimination to optimize processing speed. Further refinements include the work of Yu et al. [9], who integrated mutual information metrics to evaluate inter-feature dependencies more precisely. The field has also benefited from rough set-based methodologies, particularly the OFS-A3M algorithm [10] that utilizes neighborhood rough sets for robust feature assessment. In the domain of representation learning for incomplete datasets, latent factor analysis (LFA) has established itself as a fundamental approach. Its effectiveness stems from modeling observable data distributions while compensating for imbalance, making it particularly suitable for missing value imputation in OS²FS applications [5]. However, while LFA addresses data completeness issues, existing OSFS and OS²FS techniques still face a persistent challenge: their inability to properly handle the natural ambiguity and uncertainty present in feature boundary determination during selection processes.

## 3. Methodology

### 3.1 Phase one: missing data estimation in sparse streaming features

The proposed methodology handles sparse streaming features $B_t=\{F'_t, F'_{t+1}, …, F'_{t+H-1}\}$ through a latent factor analysis (LFA) approach. By applying a rank-L matrix approximation, the system reconstructs the complete feature representation $\hat{B}_t=\{\hat{F}'_t, \hat{F}'_{t+1}, …, \hat{F}'_{t+H-1}\}$, optimizing the solution to reduce reconstruction errors against actual observed values. The corresponding label vector space is formulated as $C=[c_1, c_2, …, c_N]^T$, where $N$ represents the dataset size and $t$ spans the time indices $\{1, 2, …, T\}$. For any individual feature component $f'_{n,j}$, the imputation process follows this computational procedure [11]-[38]:

$$\varepsilon_{n,j} = \frac{1}{2}\left(f'_{n,j} - \sum_{k=1}^{L} p_{n,k} q_{j,k}\right)^2 + \frac{\lambda}{2}\left(\sum_{k=1}^{L} p_{n,k}^2 + \sum_{k=1}^{L} q_{j,k}^2\right). \quad (1)$$

The indices in this formulation satisfy: $j\in\{t, t+1,\cdots,t+H-1\}$, $n\in\{1, 2,\cdots,N\}$, $k\in\{1, 2,\cdots,L\}$, where $\lambda$ serves as the regularization coefficient. The matrix components are constructed such that $p_{n,k}$ denotes the entry at the $n$th row and $k$th column of matrix $P$, whereas $q_{j,k}$ represents the element at the $j$th row and $k$th column of matrix $Q$. Using stochastic gradient descent (SGD) optimization, we obtain the following solution (Eq. 1):

$$\begin{aligned} p_{n,k} &\leftarrow p_{n,k} + \eta q_{j,k}\left(f'_{n,j} - \sum_{k=1}^{L} p_{n,k} q_{j,k}\right) - \lambda\eta p_{n,k}, \\ q_{j,k} &\leftarrow q_{j,k} + \eta p_{n,k}\left(f'_{n,j} - \sum_{k=1}^{L} p_{n,k} q_{j,k}\right) - \lambda\eta q_{j,k}, \end{aligned} \quad (2)$$

The parameter $\eta$ in this formulation represents the step size adjustment factor in the learning process. Through matrix factorization, the reconstructed feature space is obtained as $\hat{B}_t = PQ^T$, where $P$ and $Q$ are the factorized matrices [39]-[61].

### 3.2 Phase two: real-time evaluation of feature selection

Within our fitness-driven PSO architecture, individual particles represent potential solutions to the supervised OS²FS optimization challenge. For the m-th particle ($m\in\{1,2,...,M\}$), its personal best position $P_m=[p_{m,1}, p_{m,2}, …, p_{m,H}]$ maintains a record of its optimal local discoveries. The swarm's global best position $G=[g_1, g_2, …, g_H]$ directs collaborative search efforts. Considering a particle m characterized by its current position $X_m=[x_{m,1}, x_{m,2}, …, x_{m,H}]$ and velocity $V_m=[v_{m,1}, v_{m,2}, …, v_{m,H}]$, the evolutionary dynamics are governed by the following update equations:

$$v_{m,h}^{\gamma+1} = wv_{m,h}^{\gamma} + c_1 r_1\left(p_{m,h}^{\gamma} - x_{m,h}^{\gamma}\right) + c_2 r_2\left(g_h^{\gamma} - x_{m,h}^{\gamma}\right), \quad (3)$$

$$x_{m,h}^{\gamma+1} = x_{m,h}^{\gamma} + v_{m,h}^{\gamma+1}. \quad (4)$$

The velocity vectors of particles are modified according to Equation (3), whereas Equation (4) governs position updates using these adjusted velocities to facilitate comprehensive solution space exploration. In this formulation: $\gamma$ specifies the total number of iterations, $h\in\{1,2, …, H\}$ indexes the search space dimensions, $r_1$ and $r_2$ denote uniformly distributed random variables within [0,1], w represents the inertia weight parameter, while $c_1$ and $c_2$ correspond to cognitive and social acceleration constants (typically fixed at 2). Through successive iterations, the algorithm progressively refines particle fitness values while continuously improving both personal best (*Pbest*) and global best (*Gbest*) solutions. The complete optimization procedure executes the following sequence of operations at each iteration:

$$P_m^{\gamma+1} = \begin{cases} X_m^{\gamma+1} & \text{if } F(X_m^{\gamma+1})<F(P_m^{\gamma}), \\ P_m^{\gamma} & \text{otherwise,} \end{cases} \quad (5)$$

$$G^{\gamma+1} = \begin{cases} P_m^{\gamma+1} & \text{if } F(P_m^{\gamma+1})<F(G^{\gamma}), \\ G^{\gamma} & \text{otherwise,} \end{cases} \quad (6)$$

where the optimization process is critically governed by the fitness evaluation function $F(\cdot)$, which serves as the primary driver for solution refinement. During successive iterations, when the condition $x_{m,h}^{\gamma} > \rho$ is satisfied, the feature component $\hat{F}'_{t+h-1}$ is temporarily identified as a potential candidate. This decision-making function, expressed through the following mathematical formulation, delivers crucial quantitative evaluation for the feature selection mechanism:

$$Fitness = 1 - \frac{Correct}{N \times H}. \quad (7)$$

The evaluation metric Correct quantifies classification precision by tallying correctly classified instances based on the temporarily selected feature subset. Within this assessment paradigm, the particle achieving maximal performance is identified as representing the ideal feature combination. This outcome produces what our methodology defines as the feature evaluation matrix - a structured representation that methodically encapsulates the most distinctive features discovered through the algorithm's dynamic optimization process.

### 3.3 Phase three: dynamic three-way feature assessment

The POS²FS algorithm employs the optimized feature matrix Gb to select the most informative features. To enhance selection precision, our framework incorporates an improved three-way decision (3WD) mechanism that effectively

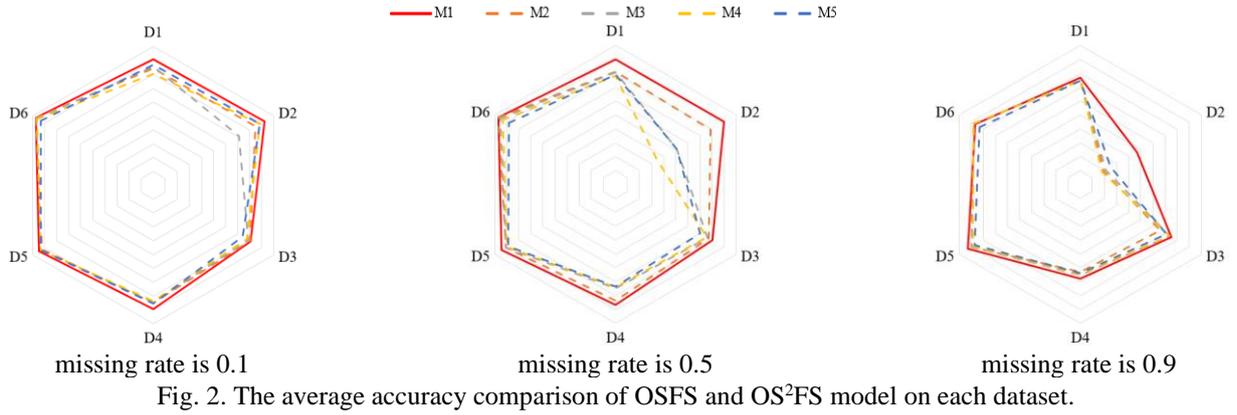

| missing rate is 0.1 | missing rate is 0.5 | missing rate is 0.9 |

Fig. 2. The average accuracy comparison of OSFS and OS²FS model on each dataset.

TABLE I THE DETAILS OF SELECTED DATASETS.

| Mark | Dataset | Features | Instances | Class |
|---|---|---|---|---|
| D1 | USPS | 242 | 1500 | 2 |
| D2 | COIL20 | 1025 | 1440 | 20 |
| D3 | Colon | 2001 | 62 | 2 |
| D4 | Lung | 3313 | 83 | 5 |
| D5 | DriveFace | 6401 | 606 | 3 |
| D6 | Lungcancer | 12534 | 181 | 2 |

TABLE II ALL THE ALGORITHM PARAMETERS.

| Mark | Algorithm | Parameter |
|---|---|---|
| D1 | UOS²FS | Z test, Alpha is 0.05, $L$=200, $w$=1, $\beta$=0.5, and $\alpha$=0.9. |
| D2 | LOSSA | Z test, Alpha is 0.05. (TSMC, 2022) |
| D3 | Fast-OSFS | Z test, Alpha is 0.05. (TPAMI, 2013) |
| D4 | SAOLA | Z test, Alpha is 0.05. (TKDD, 2016) |
| D5 | OFS-A3M | (Information Sciences, 2019) |

TABLE III ALL THE CLASSIFIER PARAMETERS.

| Classifier | Parameter |
|---|---|
| KNN | *The neighbors are 3.* |
| Random Forest | *6 decision trees.* |
| CART | *Default parameters values.* |

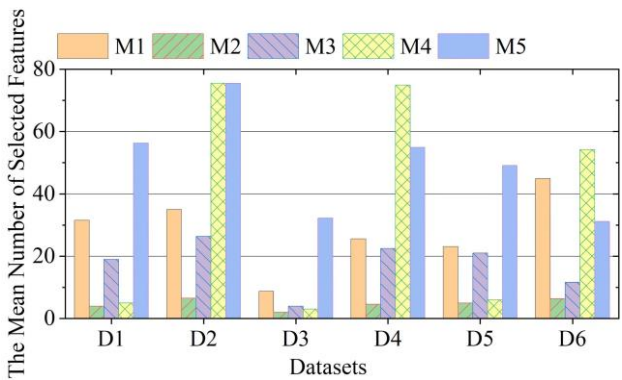

Fig. 1. The number of selected features.

TABLE IV THE RANK SUM OF THE WILCOXON SIGNED-RANKS.

| Missing rate | M2 | | M3 | | M4 | | M5 | |
|---|---|---|---|---|---|---|---|---|
| | *R+ | *R- | *R+ | *R- | *R+ | *R- | *R+ | *R- |
| 0.1 | 21 | 0 | 21 | 0 | 21 | 0 | 21 | 0 |
| 0.5 | 21 | 0 | 21 | 0 | 21 | 0 | 21 | 0 |
| 0.7 | 20 | 1 | 20 | 1 | 20 | 1 | 21 | 0 |

\* A larger value denotes a higher accuracy.

handles feature boundary uncertainty. A binary encoding scheme $\pi=\{X, X_C\}$ for feature acceptance/rejection decisions. Three action categories A=$\{a_P, a_B, a_N\}$ corresponding to positive, borderline, and negative selections. Dynamic three-region partitioning of the completed feature matrix $\hat{B}_t$ using these decision rules.

$$POS = \{\hat{F}_{t+h-1}{}' \in \hat{B}_t \,|\, gb_l \geq \alpha\},$$
$$BND = \{\hat{F}_{t+h-1}{}' \in \hat{B}_t \,|\, \beta < gb_l < \alpha\}, \quad (8)$$
$$NEG = \{\hat{F}_{t+h-1}{}' \in \hat{B}_t \,|\, gb_l \leq \beta\}.$$

The integration of three-way decision (3WD) theory provides a structured mechanism for handling feature boundary uncertainty. Our approach adopts differentiated processing strategies based on feature categorization: 1) Features classified in the *POS* (positive) region are directly incorporated into the selected feature subset $S_t$; 2) Features assigned to the *BND* (boundary) region undergo further analysis involving redundancy evaluation procedures, specifically,

$$\forall X \in S_t \text{ s.t. } P(C \,|\, \hat{F}_{t+h-1}{}', X) \neq P(C \,|\, X). \quad (9)$$

The feature component $\hat{F}'_{t+h-1}$ is identified as non-redundant and is therefore added to the final feature subset $S_t$. The elimination criteria for redundant features $\{R\}$ are defined by the following principles:

$$\forall R \in M(C)_t \cup \hat{F}_{t+h-1}{}',\, \exists \delta \subseteq M(C)_t \cup \hat{F}_{t+h-1}{}' - \{R\}$$
$$\text{s.t. } P(C \,|\, R, \delta) = P(C \,|\, \delta).$$

## 4. Results

### 4.1 General Settings

**Datasets:** We conducted extensive evaluations on six real-world datasets covering varied application domains, with detailed statistical properties provided in Table I.

**Baseline:** All benchmark methods and classification models were configured using the parameter settings documented in Tables II and III. The experiments were executed in MATLAB R2022a, running on a computing system equipped with an Intel Core i7 processor (2.40 GHz) and 16 GB memory.

### 4.2 POS²FS vs. OS²FS and OSFS Models

This study conducts a comprehensive evaluation comparing the proposed POS²FS method with standard OSFS and OS²FS techniques across ten benchmark datasets. The analysis specifically examines three distinct missing data conditions (10%, 50%, and 90% missing values). It should be noted that among the baseline methods, only LOSSA was specifically developed to handle incomplete feature streams. For other approaches (Fast-OSFS, SAOLA, and OFS-A3M), missing values were addressed through zero-imputation. The results highlight instances where performance improvements achieve statistical significance.

*4.2.1 Selected Features Analysis:* Figure 1 displays the mean number of selected features for each algorithm across datasets D1-D6. Although some baseline methods preserve almost the complete feature set, their predictive performance consistently falls below that of POS²FS. This performance gap primarily arises from suboptimal feature evaluation, leading to the retention of marginally relevant attributes. Conversely, POS²FS implements a stringent feature selection protocol that exclusively preserves highly discriminative features to optimize classification accuracy. By methodically removing non-informative and correlated features, the framework sustains reliable performance under sparse data stream conditions. This targeted feature selection approach substantially improves computational efficiency.

*4.2.2 Accuracy Analysis of Features Analysis:*
Figure 2 summarizes the overall classification accuracy of all assessed models using KNN, random forest, and CART classifiers on the benchmark datasets. We employed Wilcoxon signed-rank tests to determine the statistical significance of performance differences between our POS²FS approach and conventional methods, with comprehensive test outcomes presented in Table IV. The comparative examination of these results reveals several important findings:

a) Our experimental results reveal that the POS²FS method consistently outperforms traditional OSFS and OS²FS techniques across diverse benchmark datasets. Current OSFS implementations, which typically employ zero-imputation alongside relevance and redundancy assessments, often demonstrate two key limitations: 1) imprecise feature selection, and 2) degraded classification performance. To address these issues, our POS²FS solution innovatively combines: 1) Particle swarm optimization for efficient feature space exploration; and 2) Three-way decision making for robust feature evaluation. This dual-strategy optimization framework achieves markedly superior accuracy compared to existing approaches, as evidenced by our comprehensive experimental validation.

b) The experimental analysis reveals that LOSSA achieves optimal classification performance when handling fully sparse feature streams, confirming the effectiveness of its latent factor analysis-based completion approach. Nevertheless, the method's reliance on traditional relevance-redundancy evaluation criteria results in diminished accuracy compared to POS²FS. Our proposed POS²FS framework overcomes these constraints by incorporating: 1) Particle swarm optimization-driven feature assessment; and 2) Three-way decision-based risk-conscious selection. This combined methodology demonstrates enhanced performance in sparse streaming environments.

## 4 Conclusion

To address the limitations of existing OS²FS methods, we introduce POS²FS—a novel uncertainty-aware online sparse streaming feature selection framework leveraging particle swarm optimization. The proposed approach integrates three core innovations: 1) The LFA component adaptively completes sparse feature matrices, significantly mitigating imputation errors caused by large-scale missing data; 2) PSO-based optimization enhances feature quality assessment through guided swarm intelligence; and 3) Three-way decision theory enables adaptive feature classification with controlled risk tolerance. Comprehensive evaluations across ten benchmark datasets confirm the framework's superior performance over state-of-the-art techniques. Future work will investigate enhanced evolutionary computation paradigms for refined feature evaluation.